\documentclass[runningheads]{llncs}
\usepackage[T1]{fontenc}
\usepackage{graphicx}
\usepackage{times}
\usepackage{soul}
\usepackage{url}
\usepackage[hidelinks]{hyperref}
\usepackage[utf8]{inputenc}
\usepackage[small]{caption}
\usepackage{graphicx}
\usepackage{amsmath}
\usepackage{booktabs}
\usepackage{bbm}
\usepackage{algorithmic}
\usepackage[linesnumbered,ruled]{algorithm2e}
\usepackage[switch]{lineno}
\usepackage{bm}
\usepackage{multirow}
\usepackage[title]{appendix}
\usepackage{subcaption}
\usepackage{multicol}
\usepackage{varwidth}
\usepackage{color}
\usepackage{pifont}
\usepackage[misc]{ifsym}

\begin{document}
\title{DualMatch: Robust Semi-Supervised Learning with Dual-Level Interaction}
\renewcommand{\thefootnote}{\fnsymbol{footnote}}
\author{Cong Wang\inst{1}\footnotemark[1] \and
Xiaofeng Cao\inst{1}\footnotemark[1] (\Letter)\and Lanzhe Guo\inst{2} \and Zenglin Shi\inst{3}}

\footnotetext[1]{Equal contribution}
\authorrunning{C.Wang and X.Cao et al.}
\tocauthor{Cong~Wang, Xiaofeng~Cao, Lanzhe~Guo, and Zenglin~Shi}
\institute{School of Artificial Intelligence, Jilin University, Changchun, 130012, China\\ \email{cwang21@mails.jlu.edu.cn, xiaofengcao@jlu.edu.cn}
\and 
National Key Laboratory for Novel Software Technology, Nanjing University, Nanjing 210023, China \\ 
\email{guolz@lamda.nju.edu.cn} 
\and 
I2R, Agency for Science, Technology and Research (A*STAR)\\ 
\email{shizl@i2r.a-star.edu.sg}}

\toctitle{DualMatch: Robust Semi-Supervised Learning with Dual-Level Interaction}
\maketitle              % typeset the header of the contribution
\begin{abstract}
    Semi-supervised learning provides an expressive framework for exploiting unlabeled data when labels are insufficient. Previous semi-supervised learning methods typically match model predictions of different data-augmented views in a single-level interaction manner, which highly relies on the quality of pseudo-labels and results in semi-supervised learning not robust. In this paper, we propose a novel SSL method called DualMatch, in which the class prediction jointly invokes feature embedding in a dual-level interaction manner. DualMatch requires consistent regularizations for data augmentation, specifically, 1) ensuring that different augmented views are regulated with consistent class predictions, and 2) ensuring that different data of one class are regulated with similar feature embeddings. Extensive experiments demonstrate the effectiveness of DualMatch. In the standard SSL setting, the proposal achieves 9\% error reduction compared with SOTA methods, even in a more challenging class-imbalanced setting, the proposal can still achieve 6\% error reduction. 
    Code is available at https://github.com/CWangAI/DualMatch

\end{abstract}
\section{Introduction}

\begin{figure}[ht]
    \centering
    \includegraphics[width=0.55\textwidth]{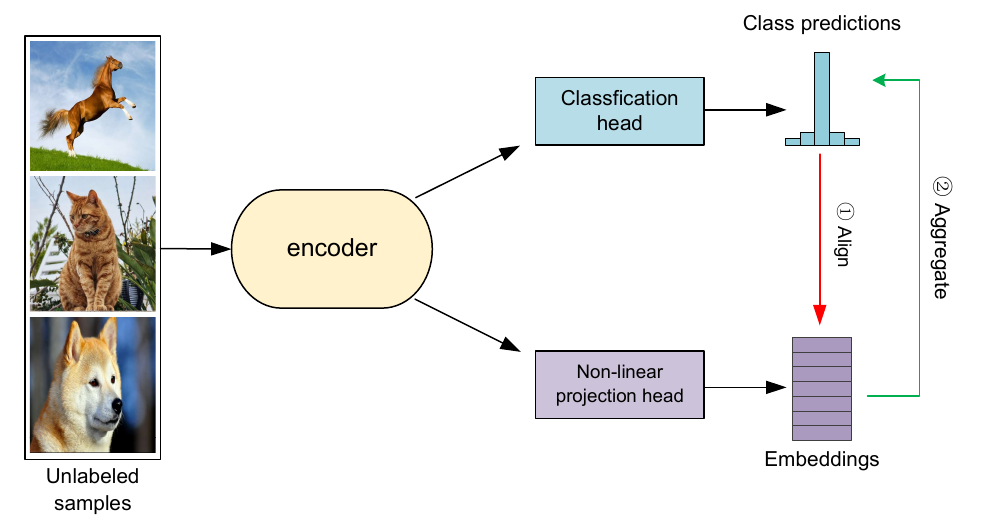}
    \caption{An illustration of DualMatch with dual-level head interaction.  Aligning the predictions of augmented data into their ground-truth labels is a single-level interaction manner (\normalsize{\textcircled{\scriptsize{1}}}) in semi-supervised learning, while the class predictions in such a manner may lack a stability guarantee for pseudo-labeling, even not robust. DualMatch  reconsiders   \normalsize{\textcircled{\scriptsize{1}}} by aligning the feature embedding of one class and then considers a  new interaction \normalsize{\textcircled{\scriptsize{2}}} by aggregating class distribution with consistent feature embeddings. 
    }
    \label{fig:1}
\end{figure}

Machine learning, especially deep learning~\cite{lecun2015deep}, has achieved great success in various tasks. These tasks, however, crucially rely on the availability of an enormous amount of labeled training data. In many real-world applications, the acquisition of labeled data is expensive and inefficient. On the contrary, there are usually massive amounts of unlabeled data. Therefore, how to exploit unlabeled data to improve learning performance is a hot topic in the machine learning community~\cite{oliver2018realistic}.

Semi-supervised learning (SSL) provides an expressive framework for leveraging unlabeled data when labels are insufficient. Existing SSL methods can be categorized into several main classes in terms of the use of unlabeled data, such as pseudo-labeling methods ~\cite{lee2013pseudo}, which assign pseudo-labels to unlabeled data based on the model prediction and train the model with labels and pseudo-labels in a supervised manner, and consistency regularization methods, which require that the output of the model should be the same when the model or data is perturbed. In much recent work, it has been reported that holistic SSL methods, e.g., MixMatch~\cite{berthelot2019mixmatch}, ReMixMatch~\cite{berthelot2019remixmatch}, and FixMatch~\cite{sohn2020fixmatch}, which consider the pseudo-labeling and consistency strategies simultaneously, have reached state-of-the-art (SOTA) performance. For example, in the image classification task, holistic SSL methods can achieve the performance of fully supervised learning even when a substantial portion of the labels in a given dataset have been discarded~\cite{sohn2020fixmatch}.

Although the holistic SSL methods have been reported to achieve positive results, they mainly adopt a single-level interaction manner between class prediction and the feature embedding, resulting in low quality of the pseudo-labels and weak SSL robustness performance. Take the SOTA FixMatch method as an example: FixMatch generates both weakly and strongly augmented views for unlabeled data, assigns high-confidence pseudo-labels predicted on the weakly augmented data to the strongly augmented one, and then optimizes the model by minimizing the cross-entropy loss between the prediction of the strongly augmented views and the corresponding pseudo-labels. This process is a single-level interaction since only different data augmentations are regulated by consistent class predictions. This results in the SSL performance being highly related to the correctness of the pseudo-label, and wrong pseudo-labels can lead to the confirmation bias of the model with error accumulation~\cite{arazo2020pseudo}. How to improve the robustness of SSL methods for pseudo-labels has emerged as a critical issue in SSL research.

In this paper, we propose a novel SSL algorithm called DualMatch. Compared with previous SSL methods that only consider the consistency between predictions for different augmentations, two consistency regularization factors are proposed in   DualMatch, which derives more robust learning performance: 1) different augmented representations of training data should be regulated with consistent class predictions, and 2) different class predictions should be regulated with consistent feature representations. We illustrate the new manner of dual-level interaction in Figure~\ref{fig:1}. Specifically, in the first-level interaction, supervised contrastive learning is utilized for aligning the feature representations of one class with highly confident predictions. This requires that the features of strongly augmented views be clustered together in the low-dimensional embedding space, and then pseudo-labels are assigned from their weakly augmented views. In the second-level interaction, class distributions with consistent feature embeddings are aggregated to generate pseudo-labels for class prediction fine-tuning. Under this dual-level learning manner, the consistency of the same data represented in two heads is enhanced, and more reliable pseudo-labels are generated for matching strongly augmented class prediction. Compared with the FixMatch algorithm, the DualMatch achieves 9\% error reduction in the CIFAR-10 dataset; even on a more challenging class-imbalanced semi-supervised learning task, the DualMatch can still achieve 6\% error reduction compared with the FixMatch algorithm.

Our contributions can be summarized as follows:
\begin{itemize}
    \item We point out that the single-level interaction that existing SSL methods commonly adopted will result in weak SSL robustness performance.
    \item We first propose the dual-level interaction between classification and feature embeddings and a novel DualMatch algorithm to improve the robustness of SSL.
    \item We rigorously evaluate the efficacy of our proposed approach by conducting experiments on standard SSL benchmarks and class-imbalanced semi-supervised learning. Our results demonstrate significant performance improvement.
\end{itemize}

\section{Related Work}

\subsection{Semi-Supervised Learning}
   A prerequisite for SSL is that the data distribution should be based on a few assumptions, including smoothness, cluster, and manifold~\cite{chapelle2006semi}. Technically, the smoothing assumption denotes that the nearby data are likely to share the same class label, and the manifold assumption denotes that the data located inner on low-dimensional streaming clusters are more likely to share the same labels. Recently, consistency regularization methods~\cite{berthelot2019mixmatch,xie2020unsupervised} have been widely applied and achieved outstanding results in SSL.  An inherent observation is that the consistency regularization could be  founded  on the manifold  or  smoothness assumption, and requires that different perturbation methods for the same data hold consistent predictions against their employed diverse models.

    From the perspective of consistency, there are two classical branches: model-level ~\cite{sajjadi2016regularization,laine2016temporal} and data-level consistency. As an early branch, ~\cite{sajjadi2016regularization} denotes the addition of random perturbation techniques (such as dropout~\cite{srivastava2014dropout} and random max-pooling methods) to the model should have consistent prediction results. To improve  its robustness, ~\cite{laine2016temporal} further aggregates the previous results of the model. 
    Considering the pseudo-label cannot vary in iterative epochs,~\cite{tarvainen2017mean} then replace the  aggregation with the exponential moving average (EMA) method. 
    Data-level consistency is established by virtual adversarial training (VAT)~\cite{miyato2018virtual} and unsupervised data augmentation (UDA)~\cite{xie2020unsupervised}. As an expressive consistency method,    VAT produces optimally augmented views by adding random noise to the data and using an adversarial attack method. Differently,   UDA utilizes the random augmentation (RA)~\cite{cubuk2020randaugment} technique to produce   strongly augmented views and minimizes the prediction disagreement between those views and their associated original data.
    Considering different levels of perturbations to the original input data, aligning different models' feedback to their early slightly perturbed inputs, i.e., anchoring,  has been proven to be more effective. A series of strategies are then presented by taking this augmentation anchoring idea. In detail, MixMatch~\cite{berthelot2019mixmatch} adopts the mixup~\cite{zhang2018mixup} trick to generate more augmented data by randomly pairing samples with each other and sharpening the average of multiple augmented data prediction distributions to generate pseudo-labels. Remixmatch~\cite{berthelot2019remixmatch} further improves the MixMatch approach by proposing a distribution alignment method (DA), which encourages the prediction of the marginal distribution of mixed data to be consistent with the true data distribution. FixMatch~\cite{sohn2020fixmatch} simply considers weakly augmented view predictions with high confidence in unlabeled data as pseudo-labels for strongly augmented views and achieves SOTA performance.

\subsection{Supervised Contrastive Learning}
   Self-supervised contrastive learning has been widely noticeable for its excellent performance by training models using unlabeled data and fine-tuning them for downstream tasks. MoCo~\cite{he2020momentum} and SimCLR~\cite{chen2020simple} establish the classical framework of self-supervised contrastive learning, which distinguishes the representations of each sample from the others. The contrastive learning frameworks consider different augmented views of the same sample as positive sample pairs and other samples as negative samples, by minimizing the info Noise Contrastive Estimation (InfoNCE) loss to pull the positive samples together and to push the negative samples away in the low-dimensional embedding space. For semi-supervised tasks, SimCLR v2~\cite{chen2020big} indicates that a big self-supervised pre-trained model is a strong semi-supervised learner and simply fine-tunes the pre-trained model by using labeled samples to train a semi-supervised model. However, self-supervised contrastive learning only considers data features without focusing on class information and causes class conflicts by pushing far away samples, resulting in the inability to be directly combined with SSL. Supervised contrastive learning~\cite{khosla2020supervised} extends the self-supervised contrastive learning methods by leveraging labeled data information to pull the samples of one class  closer and push apart clusters of samples from different classes in a low-dimensional embedding space. Therefore, supervised contrastive learning mitigates the class collision phenomenon and it can be considered for application in SSL tasks.
    
\section{Method}
    In this section, we introduce the preliminaries
    and present the two levels of interaction of  
    DualMatch. Consisting of the new    manner, the final objective   is   constructed.

\begin{figure*}[ht!]

        \centering

        \includegraphics[width=0.75\textwidth]{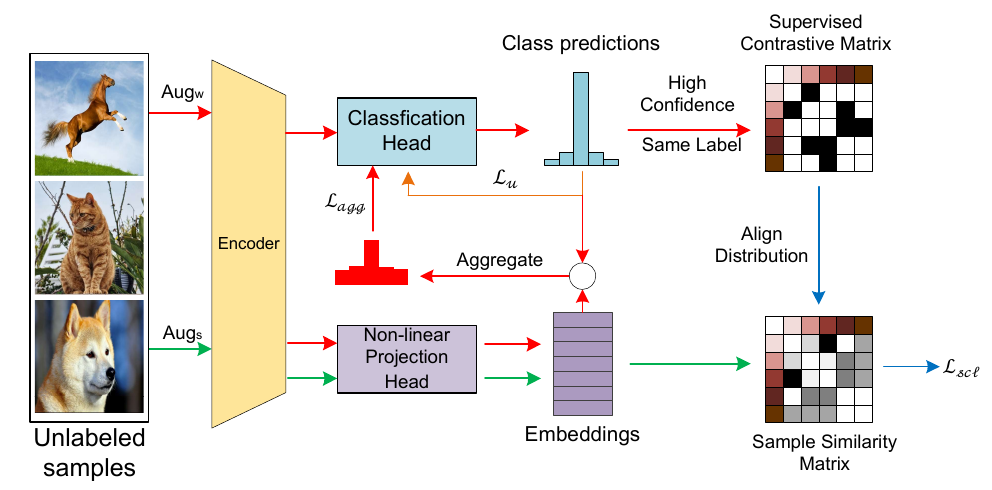}
        \caption{The framework of the proposed DualMatch. Given a batch of unlabeled images, a class prediction of weakly augmented views is generated by the classifier head. The first-level interaction aligning distribution: pseudo-labels with high confidence are used to generate the supervised contrastive matrix and the sample similarity matrix is constructed by computing the similarity between strongly augmented embeddings to match the supervised contrastive matrix. The second-level interaction aggregating pseudo-labeling: the low-dimensional embedding similarity features of the weakly augmented view are combined with predictions to aggregate the class distribution of samples. (The red and green lines indicate the process lines of the weakly and strongly augmented views, respectively.)}
        \label{fig:2}
\end{figure*}
    
\subsection{Preliminaries}
    The semi-supervised classification setting is described as follows. For an $C$-class classification problem, given a batch of $B$ labeled samples $\mathcal{X}=\left\{ \left(x_b,y_b\right): b\in\left(1,\dots,B\right)\right\}$, where $x_b$ denotes the training samples and $y_b$ denotes its one-hot label, and a batch of unlabeled samples are denoted by $\mathcal{U}=\{u_b: b\in\left(1,\dots,\mu B\right)\}$, where $\mu$  determines the relative sizes of $\mathcal{X}$ and $\mathcal{U}$. Given those settings, the next is to learn a convolutional encoder $f\left(\cdot\right)$ with labeled and unlabeled samples, a fully connected classification head $g\left(\cdot\right)$, and a non-linear projection head $h\left(\cdot\right)$. In particular, the labeled samples are randomly weakly augmented $\text{Aug}_\text{w}\left(\cdot\right)$  predicted by the classifier head $p_b=g\left(f\left(\text{Aug}_\text{w}\left(x_b\right)\right)\right)$. Then the labeled samples can be optimized with cross-entropy loss which evaluates  the ground-truth labels and the class predictions: 
    \begin{equation}\label{eq:1}
        \mathcal{L}_x = \frac{1}{B}\sum_{b=1}^{B} \text{H}\left(y_b, p_b\right),
    \end{equation}
    where $\text{H}\left(\cdot, \cdot\right)$ denotes the cross-entropy between two distributions. 
    
    Following FixMatch,~\cite{sohn2020fixmatch} apply the weak augmentation operation and the Random Augmentation method as strong augmentation operation $\text{Aug}_\text{s}\left(\cdot\right)$ to the unlabeled samples to obtain weakly and strongly augmented views respectively. The unsupervised classification loss can be defined as the cross-entropy loss of the predictions of the two views:
    \begin{equation}\label{eq:2}
         \mathcal{L}_u = \frac{1}{\mu B}\sum_{b=1}^{\mu B}\mathbbm{1}\left(\max(\text{DA}\left(p_b^w\right))\ge\tau\right) \text{H}\left(\hat{y_b}, p_b^s\right),
    \end{equation}
    where $p_b^w=g\left(f\left(\text{Aug}_\text{w}\left(u_b\right)\right)\right)$ and $p_b^s=g\left(f\left(\text{Aug}_\text{s}\left(u_b\right)\right)\right)$ refer to the prediction distributions of the weakly augmented and strongly augmented classifications of unlabeled samples. $\hat{y_b}=\text{argmax}\left(\text{DA}\left(p_b^w\right)\right)$ is the pseudo-label of the predicted weakly augmented view. $\tau$ is the pseudo-label confidence threshold. We only consider data pseudo-labels with maximum class probability greater than the threshold $\tau$. By following ~\cite{berthelot2019remixmatch}, $\text{DA}\left(\cdot\right)$ denotes the distribution alignment (DA) trick that is applied to the model's class prediction for unlabeled samples. DA maintains the predicted marginal distribution of the data consistent with the true data distribution. We compute $\Tilde{p}^w$ as the moving average of the model's predictions for unlabeled samples over the last 32 steps as the marginal distribution and adjust $p_b^w$ with $\text{Normalize}\left(p^w_b/\Tilde{p}^w\right)$. 

\subsection{The DualMatch Structure}
    \textbf{Motivation of DualMatch.} In DualMatch, we adopts dual-level interaction between class prediction $p$ of classification head and  feature embedding $z=h\left(f\left(x\right)\right)$ of non-linear projection head.
    % simultaneously consider the consistency of the class prediction  $p$ and low-dimensional feature embedding  $z=h\left(f\left(x\right)\right)$ for training, mapped by the classification head and the non-linear projection head, respectively.
    Following the augmentation anchoring method~\cite{berthelot2019remixmatch}, there are two augmented views used to represent the feature embedding, i.e., weakly augmented embedding $z^w_b$ and strongly augmented embedding $z^s_b$.  However,   some strategies  use multiple augmented views to capture the feature embedding and also obtain  promised performance. To provide a fair comparison with classical SSL methods~\cite{sohn2020fixmatch}, we generate a weakly augmented view for labeled samples and also solicit another strongly augmented view for unlabeled samples. 
    
    \noindent\textbf{Framework of DualMatch.} Figure~\ref{fig:2} illustrates the DualMatch framework with dual-level interaction. In the first-level interaction, we introduce the aligning distribution algorithm, which utilizes supervised contrastive learning to cluster the feature embedding with consistent predictions. Then, we show the aggregated pseudo-labeling method in the second-level interaction, which fine-tunes the class prediction by  aggregating pseudo-labels of similar feature embeddings.
    We below explain the two interactions of DualMatch  in detail.
    
   \subsection{First-level Interaction: Align }
   The first-level interaction aligning distribution aims to align the underlying distribution of the  class prediction and feature embedding, where its   inherent assumption is that different data of one  class should have similar feature embedding. Theoretically,  strongly augmented views  of unlabeled samples should be clustered together in the low-dimensional embedding space, while their weakly augmented views should have the same   confidence level on pseudo-labeling.   
    
    \noindent\textbf{Protocol of Aligning.} Our protocol of aligning the class prediction and feature embedding is generalized into their matrix match. Specifically, we construct a supervised contrastive matrix to solicit those predictions with high confidence from the weakly augmented views, which are required to match its associated embeddings of the sample similarity matrix from the strongly augmented views.
     
    In short, we construct the set $\mathcal{Z}= \mathcal{Z}_x\cup\mathcal{Z}_u$ of the feature embeddings, including all labeled feature embeddings and partial unlabeled feature embeddings, where  $\mathcal{Z}_x=\{(z^{x}_{b}, y_b): b\in\left(1,\dots, B\right)\}$  and $\mathcal{Z}_u=\{(z_{b}^s, \hat{y_b}):\max (p_b^w)\ge\tau, b\in\left(1,\dots,\mu B\right)\}$. Note that $\tau$ denotes the confidence level threshold. 
    
    \noindent\textbf{Supervised Contrastive Matrix} aims to obtain associations between samples from the class prediction information of the weakly augmented views. Inspired by the positive and negative sample pairs proposed by self-supervised contrastive learning~\cite{he2020momentum}, we consider the samples of one class as positive samples and samples of different classes as negative samples. In this way, we construct a supervised contrastive matrix $\mathbf{W}_{scl}$  to represent the category relationship between different samples, where the  element located at the $i$-th row and $j$-th column is defined  as follows:
    \begin{equation}\label{eq:3}
        w_{ij}^{scl}= \left\{
                \begin{array}{ll}
                  0 & ~\ \text{if} ~\ i=j ,\\
                  1 & ~\ \text{if} ~\ y_i = y_j ~\ \text{and}~\ i\neq j ,\\
                  0 & ~\ \text{otherwise} .
                \end{array}
              \right.
    \end{equation}
 % In the second stage of DualMatch,  we aggregate the pseudo-labels of similar 
%feature embeddings to fine-tune 
%the class  predictions.
    \begin{remark}
        Following contrastive learning, each sample is used as an anchor for the other samples, not as a positive sample. We thus set  the samples with the same indices (i.e., elements on the diagonal) as 0 and the samples with the same labels as 1. 
    \end{remark}
    \noindent\textbf{Sample Similarity Matrix} aims to obtain the similarity between the low-dimensional feature embeddings of the samples. The sample similarity matrix $\mathbf{S}$ is constructed by computing the similarity between embeddings in the set $\mathcal{Z}$. For each element $s_{ij} \in \mathbf{S}$, it is characterized by the cosine similarity, i.e.,  
    \begin{equation}\label{eq:4}
        {\rm sim}(z_i, z_j) = \frac{z_i^Tz_j}{\left\|z_i\right\|\left\|z_j\right\|}.
    \end{equation}
    where $z_i$ and $z_j$ are feature embeddings of   $\mathcal{Z}$.

    Recalling the protocol of aligning, improving the consistency between  the class predictions and   feature embeddings  can be achieved by matching two matrices $\mathbf{W}_{scl}$ and $\mathbf{S}$. Due to the disagreement of metrics in the two matrices, we employ  the InfoNCE loss of supervised contrastive learning~\cite{khosla2020supervised} to align their elements:
    \begin{equation}\label{eq:5}
    \begin{split}
	\mathcal{L}_{scl}=&\sum_{i \in I}\mathcal{L}_i(z_i)\\
	=&\sum_{i \in I} \frac{-1}{|\mathcal{J}(i)|} \sum_{j \in \mathcal{J}(i)} \log \frac{\exp \left(z_i \cdot z_j / t\right)}{\sum_{a \in A(i)} \exp \left(z_i \cdot z_a / t\right)},
    \end{split}
    \end{equation}
    where $i \in I$ denotes the indices of the embedding in $Z$, $A(i)= I\backslash \left\{i\right\}$ denotes the set of indices without $i$,  $\mathcal{J}(i)=\left\{j \in A\left(i\right):y_j=y_i\right\}$ is the indices of the set of positive instances of the same label as $i$, and $t$ is a temperature parameter. Let $\mathbf{W}_i$ denote the $i$-th row of the matrix $\mathbf{W}_{scl}$. To facilitate computer calculations, the Eq.~(\ref{eq:5}) can be simplified by the elements in matrix $\mathbf{W}_{scl}$ and $\mathbf{S}$ as follows:
    \begin{equation}\label{eq:6}
        \mathcal{L}_{scl}=-\sum_{i \in I}\frac{1}{\|\mathbf{W}_i\|}\sum_{j \in I}\log\frac{w_{ij}^{scl}\cdot \exp\left(s_{ij}/t\right)}{\sum_{a \in A(i)}\exp\left( s_{ia}/t\right)}
    \end{equation}

    \subsection{Second-level Interaction: Aggregate  } 
  The  second-level interaction aggregating pseudo-labeling aims to aggregate class distributions with consistent feature embeddings to generate pseudo-labels for class prediction fine-tuning. Intuitively, samples with similar features embeddings in the low-dimensional embedding space should have the same labels, so that, for a batch of unlabeled samples, we can generate aggregated pseudo-labels by aggregating the class predictions of each sample's neighbors in the embedding space to improve pseudo-labeling robustness. To avoid the cumulative error caused by the class predictions of dissimilar samples, we select $K$ neighbor samples with the most similar feature embeddings. Then the aggregated pseudo-label $q^{w}_b$ of  $u_b$ in a batch of   unlabeled samples can be defined as follows:
    \begin{equation}\label{eq:7}
        q^{w}_b = \frac{1}{K}\sum_{k=1}^{K}{\rm sim}(z_b^{w}, z_k^{w})\cdot p_k^w,
    \end{equation}
    where $p^w_k$ and $z^w_k$ denote the class prediction and feature embedding of weakly augmented unlabeled views, respectively. In particular, the class distribution is weighted by the similarity ${\rm sim}(z_b^{w}, z_k^{w})$ of the samples to their neighbors. Since the weighted class distribution cannot directly represent the classification probabilities, we adjust $q_b^w = \text{Normalize}\left(q_b^w\right)$. Like the unsupervised classification loss, we only consider the samples with high confidence aggregated pseudo-labels. The difference is that aggregated pseudo-label is soft~(a vector of probabilities) because we aim to adjust the class predictions. The aggregation loss can be optimized by cross-entropy as follows:
    \begin{equation}\label{eq:8}
        \mathcal{L}_{agg}=\frac{1}{\mu B}\sum_{b=1}^{\mu B}\mathbbm{1}\left(\max\left(q_b^w\right)\ge\tau_1\right) \text{H}\left(q_b^w, p_b^s\right),
    \end{equation}
    where $\tau_1$ is the confidence threshold of the aggregated label. 
    
    \subsection{Final Objective}   The overall loss of the semi-supervised DualMatch method consists of the supervised loss $\mathcal{L}_{x}$ (w.r.t. Eq.~(\ref{eq:1})) and unsupervised loss $\mathcal{L}_{u}$ (w.r.t. Eq.~(\ref{eq:2})). Meanwhile, to   achieve the consistency of the classification prediction   and feature embedding,  we add the supervised contrastive loss $\mathcal{L}_{scl}$ (w.r.t. Eq.~(\ref{eq:6})), and aggregation loss $\mathcal{L}_{agg}$ (w.r.t. Eq.~(\ref{eq:8})). In such settings, our optimization objective is to minimize the overall loss:
    \begin{equation}\label{eq:9}
    \mathcal{L}_{overall}=\mathcal{L}_x+\lambda_u\mathcal{L}_u+\lambda_{scl}\mathcal{L}_{scl}+\lambda_{agg}\mathcal{L}_{agg},
    \end{equation}
    where $\lambda_u$, $\lambda_{scl}$, and $\lambda_{agg}$ are hyperparameters used to control the weights of loss. DualMatch can be summarized as Algorithm~\ref{alg}.

\begin{algorithm}[ht]

\caption{DualMatch algorithm.}
\label{alg}

	\textbf{Input:} Labeled batch $\mathcal{X}=\left\{ \left(x_b,y_b\right): b\in\left(1,\dots,B\right)\right\}$, unlabeled batch $\mathcal{U}=\{u_b: b\in\left(1,\dots,\mu B\right)\}$, encoder $f(\cdot)$, classification head $g(\cdot)$, non-linear projection head $h(\cdot)$.\\
    \For{step=1 to total-step}{
        $p_b=g\left(f\left(\text{Aug}_\text{w}\left(x_b\right)\right)\right)$\quad
        $z_b^x=h\left(f\left(\text{Aug}_\text{w}\left(x_b\right)\right)\right)$
        
	   $p_b^w=g\left(f\left(\text{Aug}_\text{w}\left(u_b\right)\right)\right)$ \quad
            $z_b^w=h\left(f\left(\text{Aug}_\text{w}\left(x_b\right)\right)\right)$\\
            $p_b^s=g\left(f\left(\text{Aug}_\text{s}\left(u_b\right)\right)\right)$\quad
            $z_b^s=h\left(f\left(\text{Aug}_\text{s}\left(x_b\right)\right)\right)$\\
            $\hat{y_b}=\text{argmax}\left(\text{DA}\left(p_b^w\right)\right)$\\
  
        Construct feature embedding set $\mathcal{Z}= \mathcal{Z}_x\cup\mathcal{Z}_u$\\
         $\mathcal{Z}_x=\{(z^{x}_{b}, y_b): b\in\left(1,\dots, B\right)\}$ \\$\mathcal{Z}_u=\{(z_{b}^s, \hat{y_b}):\max (p_b^w)>\tau, b\in\left(1,\dots,\mu B\right)\}$\\
	\For {$i \in \{1,...,\mu B\}$ and $j \in \{1,...,\mu B\}$}	
	{

               $w_{ij}^{scl}= \in \mathbf{W}_{scl}$ is constructed by Eq.(~\ref{eq:3})
              
               $s_{ij} \in \mathbf{S}$ is constructed by Eq.~(\ref{eq:4})
		 
}
       
            $q^{w}_b = \frac{1}{K}\sum_{k=1}^{K}{\rm sim}(z_b^{w}, z_k^{w})\cdot p_k^w$\\
            $q_b^w = \text{Normalize}\left(q_b^w\right)$

	$\mathcal{L}_x = \frac{1}{B}\sum_{b=1}^{B} \text{H}\left(y_b, p_b\right),$\\
	$\mathcal{L}_u = \frac{1}{\mu B}\sum_{b=1}^{\mu B}\mathbbm{1}\left(\max(\text{DA}\left(p_b^w\right))\ge\tau\right) \text{H}\left(\hat{y_b}, p_b^s\right)$\\
	$\mathcal{L}_{scl}=-\sum_{i \in I}\frac{1}{\|\mathbf{W}_i\|}\sum_{j \in I}\log\frac{w_{ij}^{scl}\cdot \exp\left(s_{ij}/t\right)}{\sum_{a \in A(i)}\exp\left( s_{ia}/t\right)}$ \\
    $\mathcal{L}_{agg}=\frac{1}{\mu B}\sum_{b=1}^{\mu B}\mathbbm{1}\left(\max\left(q_b^w\right)\ge\tau_1\right) \text{H}\left(q_b^w, p_b^s\right)$\\
	$\mathcal{L}_{overall}=\mathcal{L}_x+\lambda_u\mathcal{L}_u+\lambda_{scl}\mathcal{L}_{scl}+\lambda_{agg}\mathcal{L}_{agg}$	\\
    Optimize $f(\cdot)$, $g(\cdot)$ ,and $h(\cdot)$ by minimizing $\mathcal{L}_{overall}$}
	
    \textbf{Output:} Trained model.

\end{algorithm}

   \noindent\textbf{Exponential Moving Average.}
    From the perspective of consistent model regularization, we employ the  Exponential Moving Average (EMA) strategy    ~\cite{tarvainen2017mean} to
    smooth the model parameters  with an expectation of lower variation. Technically,  the parameters of EMA  are usually weighted by   previously associated model parameters in the iterative updates:
    \begin{equation}
        \overline{\theta}=m\overline{\theta}+\left(1-m\right)\theta,
    \end{equation}
    where $\overline{\theta}$ denotes the parameters of the EMA model, $\theta$ denotes the parameters of the training model, and $m$ denotes the EMA decay rate. Note that the experiments also employ  the EMA model for testing.

\section{Experiment}

    In this section, we evaluate DualMatch on several semi-supervised tasks including  semi-supervised classification and class-imbalanced semi-supervised classification. Our ablation studies the effect of dual-level interaction and hyperparameters on the framework.

   \begin{table*}[t]

    \centering
    \caption{Error rate (mean$\pm$std \%) of semi-supervised classification of DualMatch vs. baseline methods  over varying numbers of labeled samples (5 runs). }
    \setlength\tabcolsep{8pt}
    \begin{center}
    \resizebox{0.9\textwidth}{!}{%S
    \begin{tabular}{lrrrrrr}
       \toprule
        & \multicolumn{3}{c}{CIFAR-10} & \multicolumn{2}{c}{CIFAR-100} & \multicolumn{1}{c}{STL-10} \\
        \cmidrule(l{3pt}r{3pt}){2-4}  \cmidrule(l{3pt}r{3pt}){5-6} \cmidrule(l{3pt}r{3pt}){7-7}
        Method & 40 labels & 250 labels & 4000 labels & 2500 labels & 10000 labels & 1000 labels \\
        \cmidrule(l{3pt}r{3pt}){1-1} \cmidrule(l{3pt}r{3pt}){2-4}  \cmidrule(l{3pt}r{3pt}){5-6} \cmidrule(l{3pt}r{3pt}){7-7}
        $\Pi$-Model      &  74.34$\pm$1.76 &  54.26$\pm$3.97   &  41.01$\pm$0.38 &  57.25$\pm$0.48   &  37.88$\pm$0.11 & 32.78$\pm$0.40\\
        Pseudo-Labeling  &  74.61$\pm$0.26 &  49.78$\pm$0.43   &  16.09$\pm$0.28 &   57.38$\pm$0.46   &  36.21$\pm$0.19 & 32.64$\pm$0.71\\
        Mean Teacher     &  70.09$\pm$1.60 &  32.32$\pm$2.30   &  9.19$\pm$0.19 &   53.91$\pm$0.57   &   35.83$\pm$0.24 & 33.90$\pm$1.37\\
        MixMatch         &  47.54$\pm$11.50&  11.05$\pm$0.86   &  6.42$\pm$0.10 & 39.94$\pm$0.37   & 28.31$\pm$0.33 & 21.70$\pm$0.68\\
        UDA              &  29.05$\pm$5.93 &  8.82$\pm$1.08   &  4.88$\pm$0.18 & 33.13$\pm$0.22  &24.50$\pm$0.25 & 6.64$\pm$0.17\\
        ReMixMatch       &  19.10$\pm$9.64 &  5.44$\pm$0.05   &   4.72$\pm$0.13 & 27.43$\pm$0.31   & 23.03$\pm$0.56 & 	6.74$\pm$0.14\\
        FixMatch         &  13.81$\pm$3.37 &  5.07$\pm$0.65   &  4.26$\pm$0.05 & 28.29$\pm$0.11  & 22.60$\pm$0.12 & 6.25$\pm$0.33\\
        CoMatch          &  6.91$\pm$1.39 &  4.91$\pm$0.33   &  4.06$\pm$0.03   & 27.18$\pm$0.21  & 21.83$\pm$0.23 & 8.66$\pm$0.41\\
        CR          &  \textbf{5.69}$\pm$0.90    &  5.04$\pm$0.30   &  4.16$\pm$0.13  & 27.58$\pm$0.37  & 21.03$\pm$0.23 & 6.96$\pm$0.42\\
        
        \cmidrule(l{3pt}r{3pt}){1-1} \cmidrule(l{3pt}r{3pt}){2-4}  \cmidrule(l{3pt}r{3pt}){5-6} \cmidrule(l{3pt}r{3pt}){7-7}
        
        DualMatch(Ours)  & 5.75$\pm$1.01 & \textbf{4.89}$\pm$0.52  & \textbf{3.88}$\pm$0.10 & \textbf{27.08}$\pm$0.23 & \textbf{20.78}$\pm$0.15 & \textbf{5.94} $\pm$0.08 \\
        \bottomrule
    \end{tabular}
    }%
    \end{center}
    
    \label{table:1-cifar_result}

\end{table*}

\subsection{Semi-supervised  Classification }
    First, we evaluate DualMatch on the semi-supervised  classification using the CIFAR-10, CIFAR-100 and STL-10 datasets. CIFAR-10 consists of 60,000 32$\times$32 images divided into 10 classes, with 6,000 images in each class. There are 50,000 training images and 10,000 test images. Following the widely adopted setting in SSL studies Fixmatch~\cite{sohn2020fixmatch}, we randomly select 4, 25, and 400 samples per class from the training set as labeled data and then use the rest of the training set as unlabeled data, respectively. In this setting, CIFAR-100 has the same number of training set and test set images as CIFAR-10, while CIFAR-100 is divided into 100 classes with 600 images in each class. We thus randomly select 25, 100 samples per class as labeled data. STL-10 has 5,000 labeled and 100,000 unlabeled 96$\times$96 images in 10 classes for training, and 8,000 images for testing. We randomly select 100 samples per class from labeled images as labeled data. Please note that we evaluate the experiment with different random seeds for 5 runs.

    \noindent\textbf{Implementation Details.} We use the Wide ResNet-28-2~\cite{zagoruyko2016wide} with a weight decay of 0.0005 for the CIFAR-10, Wide ResNet-28-8 with a weight decay of 0.001 for the CIFAR-100 , and Wide ResNet-37-2 with a weight decay of 0.0005 for the STL-10. The classification head is a softmax layer and the non-linear projection head is set as a two-layer MLP. Following the implementation of~\cite{sohn2020fixmatch}, the model uses the SGD optimizer with the Nesterov momentum~\cite{sutskever2013importance} of 0.9. For the learning rate, we use the cosine learning rate decay and set the learning rate to $0.03\cdot \cos\left(\frac{7\pi n}{16N}\right)$, where $n$ denotes the current training steps and $N$ denotes the number of the total training steps. For the rest of the hyperparameters, we set $\lambda_u=1$, $\lambda_{scl}=1$, $\lambda_{agg}=0.5$, $\mu=7$, $B=64$, $\tau=0.95$, $\tau_1=0.9$, and $t=0.5$, $m=0.999$. For the training steps, we set $N=2^{20}$  for CIFAR-10, STL-10 and $N=2^{19}$ for CIFAR-100. Moreover, we utilize the warm-up trick to train aggregation loss after the first $30\times2^{10}$ training steps. For the  neighbor settings, we set $K=10$ for  CIFAR-10, STL-10 and $K=2$ for CIFAR-100. For data augmentation, we follow the implementation details of the weak and strong augmentation of FixMatch~\cite{sohn2020fixmatch}.

    \noindent\textbf{Compared Methods.} We compare with the following baseline methods: 1) Model-level consistency methods including the $\Pi$-Model~\cite{rasmus2015semi}, Pseudo-labeling~\cite{lee2013pseudo}, and Mean Teacher~\cite{tarvainen2017mean}, 2) Data-level methods including the UDA~\cite{xie2020unsupervised}, MixMatch~\cite{berthelot2019mixmatch}, ReMixMatch~\cite{berthelot2019remixmatch}, FixMatch~\cite{sohn2020fixmatch}, CoMatch~\cite{li2021comatch},  CR~\cite{lee2022contrastive}.
    
    \noindent\textbf{Results.} The SSL results are presented in Table~\ref{table:1-cifar_result}, where DualMatch achieves SOTA performance at different number settings of labeled samples. For  model-level consistency, we observe that  the $\Pi$-model, Pseudo-Labeling, and Mean Teacher perform  poorly with extremely few numbers of labeled samples, but the improvement in error rate becomes more significant after adding more labeled samples. It is thus the model-level consistency semi-supervised methods that are highly dependent on the number of labeled samples.  For data-level  consistency, we observed that     the performance of UDA, MixMatch, ReMixMatch, and FixMatch with the help of data augmentation methods improved significantly in extremely few labeled samples. Moreover, the performance of the semi-supervised methods using strong augmentation (e.g., Randaugment~\cite{cubuk2020randaugment}) tricks exceeds that of simple that of the  simple  tricks for  data augmentation, e.g., mixup. 
    Therefore, the data-level consistency semi-supervised methods utilize various data augmentation tricks to overcome the shortcoming of insufficient labeled data volume. Compared to the above methods,  CoMatch and DualMatch  introduce feature embedding to further exploit the underlying distribution of classes, and the error rate reduction of training on 40 labeled samples of CIFAR-10 is much better than that of the data-level and model-level  consistency methods. Furthermore, training on 250 and 4000 labeled samples of CIFAR-10 also achieves attractive  results, but not so significantly as 40 labeled samples. Additionally, compared with  FixMatch,  DualMatch achieves a 9\% error reduction in  CIFAR-10. The potential result is that such a semi-supervised training manner with efficient feature embeddings  performs closely to fully supervised training in CIFAR-10.

\subsection{Class-imbalanced Semi-supervised Classification}
 
\begin{table}[t]

    \centering
    
    \caption{Error rate (mean$\pm$std \%) for CIFAR-10 with the labeled ratio $\beta=10\%$  and  imbalance ratio $\gamma=\{50, 100, 200\}$ (5 runs).} 
    \label{tab:2-cifar10-baselines}
    \setlength\tabcolsep{8pt}
    \begin{tabular}{lrrr}
    \toprule
        Method          &    $\gamma=50$   & $\gamma=100$  & $\gamma=200$ \\
         \cmidrule(l{3pt}r{3pt}){1-1}  \cmidrule(l{3pt}r{3pt}){2-4}
        Pseudo-Labeling &    47.5$\pm$0.74 & 53.5$\pm$1.29 & 58.0$\pm$1.39 \\
        Mean Teacher    &    42.9$\pm$3.00 & 51.9$\pm$0.71 & 54.9$\pm$1.28 \\
        MixMatch        &    30.9$\pm$1.18 & 39.6$\pm$2.24 & 45.5$\pm$1.87 \\
        FixMatch      &    20.6$\pm$0.65 & 33.7$\pm$1.74 & 40.3$\pm$0.74 \\
        FixMatch w/ DA    &    19.8$\pm$0.45 & 30.3$\pm$1.27 & 38.0$\pm$0.84 \\
        CoMatch   &    19.7$\pm$0.68 & 28.6$\pm$1.85 & 40.0$\pm$1.56 \\
        \cmidrule(l{3pt}r{3pt}){1-1}  \cmidrule(l{3pt}r{3pt}){2-4}
        DualMatch(Ours)   &    \textbf{19.0}$\pm$0.82 & \textbf{28.3}$\pm$1.38  & \textbf{37.3}$\pm$0.39  \\
        \bottomrule
    \end{tabular}

\end{table}

    Standard SSL assumes the class distribution is balanced, however, in real-world tasks, the data distribution is often class-imbalanced~\cite{mahajan2018exploring}. How to guarantee the performance robustness of SSL algorithms under class-imbalanced settings is an important problem that has attracted the great attention of SSL researchers~\cite{kim2020distribution,wei2021crest}. Therefore, we also conduct experiments to evaluate the effectiveness of our proposal on class-imbalanced semi-supervised learning problems.
    % Semi-supervised classification on class-imbalanced data is a realistic problem, class-imbalanced data further challenges the task of SSL with fewer labeled samples. 
    DARP~\cite{kim2020distribution} denotes that class-imbalanced data biases SSL methods in generating pseudo-labels for the majority classes. To evaluate the effectiveness of the semi-supervised model in the class-imbalance task, we compare the results of Dualmatch and major semi-supervised methods under imbalanced data distribution.
    
    \noindent\textbf{Problem Setup.} By following ~\cite{wei2021crest}, for an $C$-class classification problem, given a labeled set $\mathcal{X} = {(x_m, y_m) : m \in \left(1, \dots , M\right)}$, where $x_m$ are the training samples and $y_m$ are one-hot labels. The number of class $c$ in $\mathcal{X}$ is denoted as $M_c$ and $\sum_{c=1}^{C}M_c=M$. ~\cite{wei2021crest} assume that the marginal class distribution of $\mathcal{X}$is skewed and the classes are ordered by decreasing order, i.e. $M_1\geq M_2\geq\cdots\geq M_C $. Class imbalance can be measured by the imbalance ratio $\gamma = \frac{M_1}{M_C}$. And given a unlabeled set $\mathcal{U} = {(u_l : l \in \left(1, \dots, L\right)}$ with the same class distribution as $\mathcal{X}$. The labeled ratio $\beta=\frac{M}{M+L}$ denotes the percentage of labeled data to the training data. Specifically, the CIFAR-10 dataset consists of 5000 images in each class, and the imbalanced majority class employs 5000 images. The setting of our evaluation experiment is on CIFAR-10 with the labeled ratio $\beta$ of 10\%, i.e. 500 labeled images and 4500 unlabeled images in the majority class and the imbalance ratio $\gamma$ of 50, 100, and 200, respectively. For the evaluation criterion of the experiment, the data of the test set is class-balanced. %without any changes.

    \noindent\textbf{Implementation Details.} We use mostly the same parameter settings as for the semi-supervised classification task, except that the number of neighbor samples $K$ is set to 2. For each experimental setting, the training steps are set to $2^{17}$ for MixMatch and $2^{16}$ for FixMatch and CoMatch. For a fair comparison, we set the total training steps to $2^{16}$ for DualMatch. For each experiment, we evaluate 5 times with different random seeds and report the mean and std of the test error rate. We report the performance using the EMA model.

    \noindent\textbf{Results.} The results of class-imbalanced semi-supervised classification are presented in Table~\ref{tab:2-cifar10-baselines}. Overall,  the DualMatch achieves better performance than the typical semi-supervised baselines using different imbalance ratios. Moreover,  all semi-supervised baselines are affected by class-imbalanced data, and their error rate increases with the increase of the imbalance ratio. For this ratio, we also observe that the data-level consistency baselines achieve the best performance if
    the imbalance ratio is set as 100, at least better than the setting of 50 and 200.  The potential reasons are as follows. For the imbalance ratio of 200, there is only 1 labeled sample for the minority class, which leads difficult to learn the features of the minority class during model training. For the imbalance ratio of 50, the effect of imbalanced data is not significant in causing class bias, but rather in the increase in error rate due to the reduction of training samples. For the imbalance ratio of 100, the class bias caused by class-imbalanced data leads to instability of the model and increases the std of error rate. The results show that CoMatch is more affected by the imbalance ratio and performs poorly at the imbalance ratio of 200, and the improvement of DualMatch is effective. We can conclude that DualMatch aligns the feature embeddings of one class during the training period, which can separate the features of different classes and make the classification boundary clearer. It also adjusts the bias of class prediction by aggregating feature embeddings to enhance the robustness of the classification boundary and mitigate the influence of class with few samples from others. Additionally,  DualMatch achieves a 6\% error reduction at the imbalance rate of 100 compared to FixMatch, and a 6.5\% error reduction at the imbalance rate of 200 compared to CoMatch.
    
    \begin{figure*}[t]

    \centering
    
    \subcaptionbox{High-confidence Sample Ratio \label{fig:high_ratio}}{\includegraphics[width=.32\textwidth]{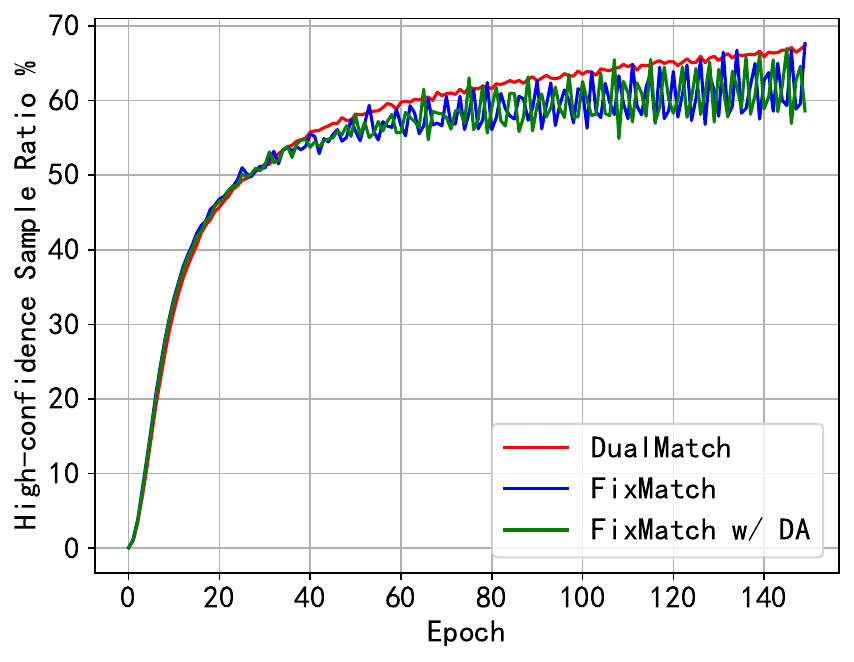}}
    \subcaptionbox{Unlabeled Sample Error Rate \label{fig:unlabel_err}}{\includegraphics[width=.32\textwidth]{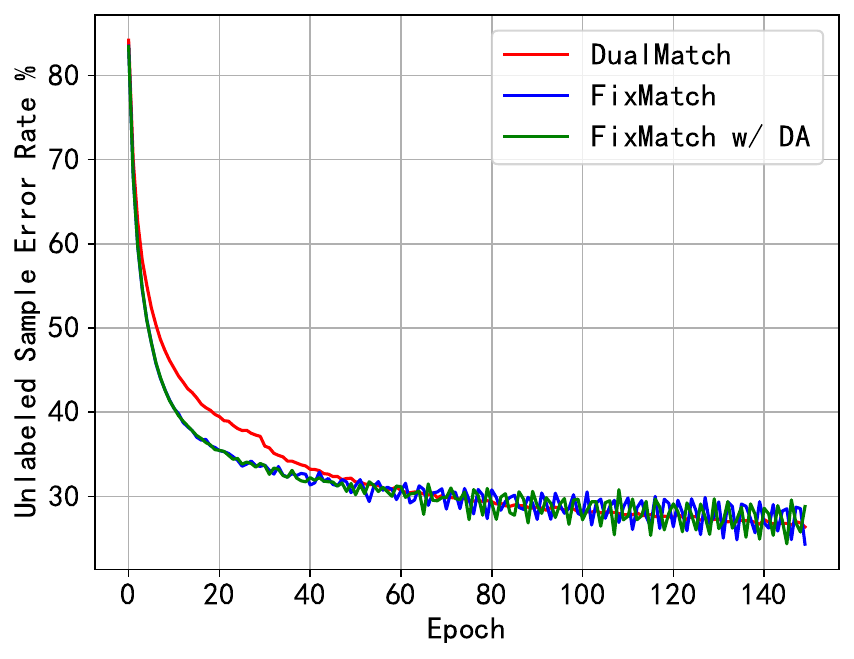}}
    \subcaptionbox{Test Error Rate \label{fig:test_err}}{\includegraphics[width=.32\textwidth]{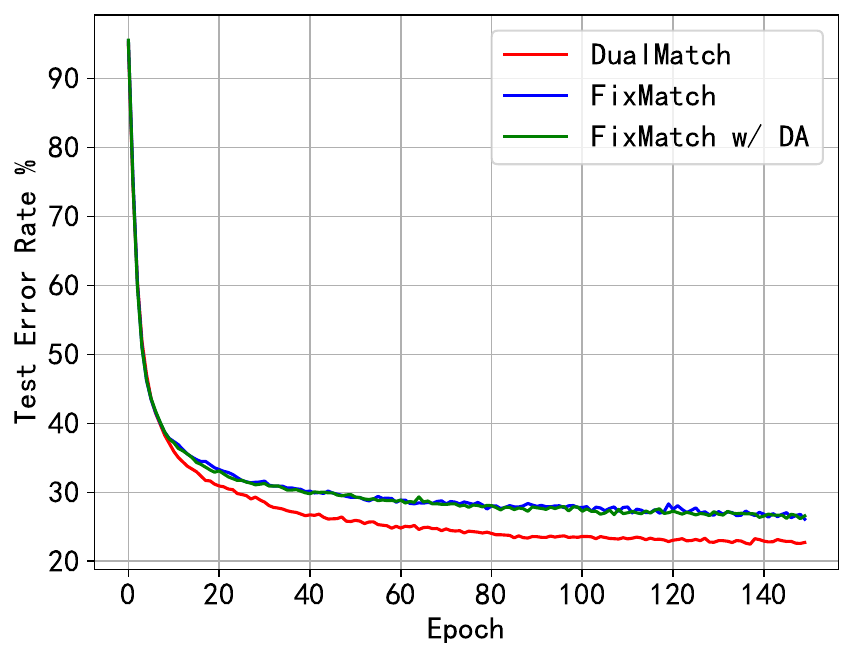}}

    \caption{The training process of FixMatch and DualMatch on CIFAR-100 with 10000 labeled samples. (a) Ratio of samples with high-confidence pseudo-labels. (b) Error rate of all unlabeled sample. (c) Error rate of test samples.}
    \label{fig:train_process}
\end{figure*}

\subsection{Ablation Study}

We study the unlabeled data error rate of FixMatch and DualMatch on the setting of  training CIFAR-100 with 10000 labeled samples. This helps us to reveal the potential influence of pseudo-labeling on semi-supervised training.
Then, we analyze the head interaction and parameter perturbation of each level of DualMatch on the setting of  training CIFAR-10 with 250 labeled samples.

    \noindent\textbf{Unlabeled Samples Error Rate.} In Figure~\ref{fig:train_process}, we study the training process of FixMatch, FixMatch with DA, and DualMatch on the setting of training CIFAR-100 with 10000 labeled samples. The potential observation factors  are 1) the  unlabeled sample error rate, and 2) the sample ratio of high-confidence  pseudo-labels. 
    From the presented curves of Figures~\ref{fig:unlabel_err} and \ref{fig:high_ratio}, as the number of training epochs increases, both the unlabeled sample error rate and high-confidence sample ratio of FixMatch fluctuate dramatically and become increasingly unstable. It is worth noting that the DualMatch starts with a high unlabeled sample error rate in the first few epochs, however, as the number of training epochs increases, the unlabeled sample error rate decreases more smoothly to the FixMatch level. The DualMatch achieves a lower test error rate than FixMatch and FixMatch with DA throughout the training process. The results show that the pseudo-labels of the unlabeled samples of FixMatch vary continuously,  which makes the learning model's poor stability even worse and affects the classification results.  In contrast, DualMatch provides more robust and high-quality pseudo-labeling during training, which significantly improves the performance of the semi-supervised learning model.

    \noindent\textbf{Align Distribution.} We
    vary the labeled and unlabeled augmentation views of the feature embedding set to perform the ablation study of Align Distribution (AD).
   The results are presented in Table~\ref{table:Abaltion-scl}. Note that when the feature embeddings are not used, the experiments fall back to DualMatch without AD. Furthermore, we also observe that simultaneously employing both labeled and unlabeled feature embeddings can effectively improve the model performance.
    \begin{table}[t]
    \centering
    \setlength\tabcolsep{6pt}
    \caption{Error rate(\%) of varying the labeled and unlabeled augmentation views of the feature embedding set.}
    \label{table:Abaltion-scl}
    \begin{tabular}{l|r|r|r|r}
       \toprule  
       \multirow{2}{*}{Ablation}& \multicolumn{1}{|c|}{Labeled} & \multicolumn{2}{c|}{Unlabeled} & \multirow{2}{*}{Error Rate} \\
         & \multicolumn{1}{c}{Weak} & \multicolumn{1}{|c}{Weak} & \multicolumn{1}{c|}{Strong} &  \\

        \cmidrule(l{3pt}r{3pt}){1-5}
        DualMatch      &  1 &  0 &  1 &   4.41\\
        w/o AD         &  0 &  0 &  0 &   4.77\\
        w/o AD ($\tau_1=0.6$)        &  0 &  0 &  0 &   5.49\\
        w/o labeled    &  0 &  0 &  1 &   4.68\\
        w/o unlabeled  &  1 &  0 &  0 &   4.93\\ 
        w/  multi      &  2 &  1 &  1 &   4.48\\
        \bottomrule
    \end{tabular}

\end{table}

\begin{figure}[!t]

    \centering
    \subcaptionbox{Number of Neighbors. \label{fig:neighbor}}{\includegraphics[width=.32\textwidth]{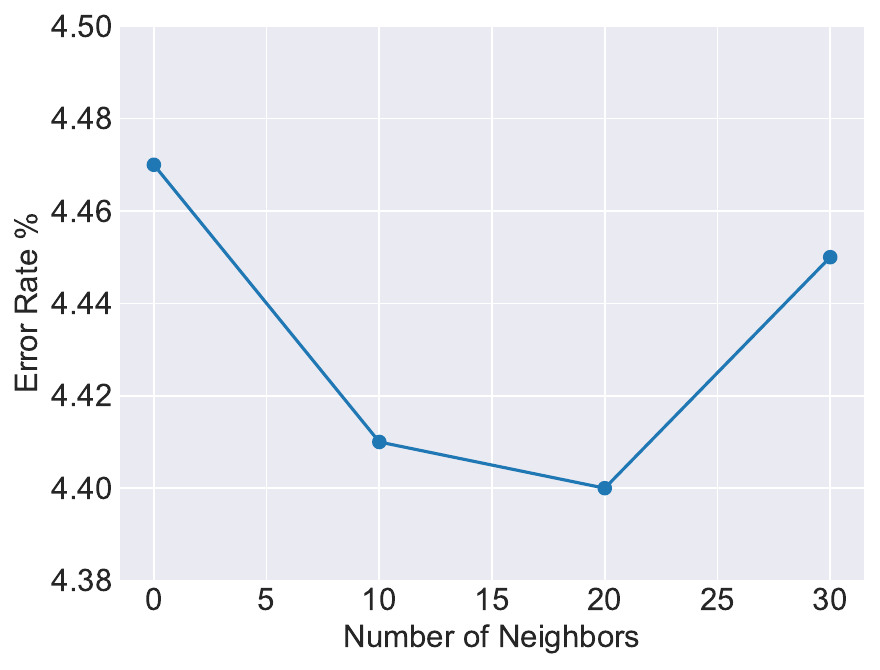}}
    \subcaptionbox{Aggregation Threshold. \label{fig:thershoud_1}}{\includegraphics[width=.32\textwidth]{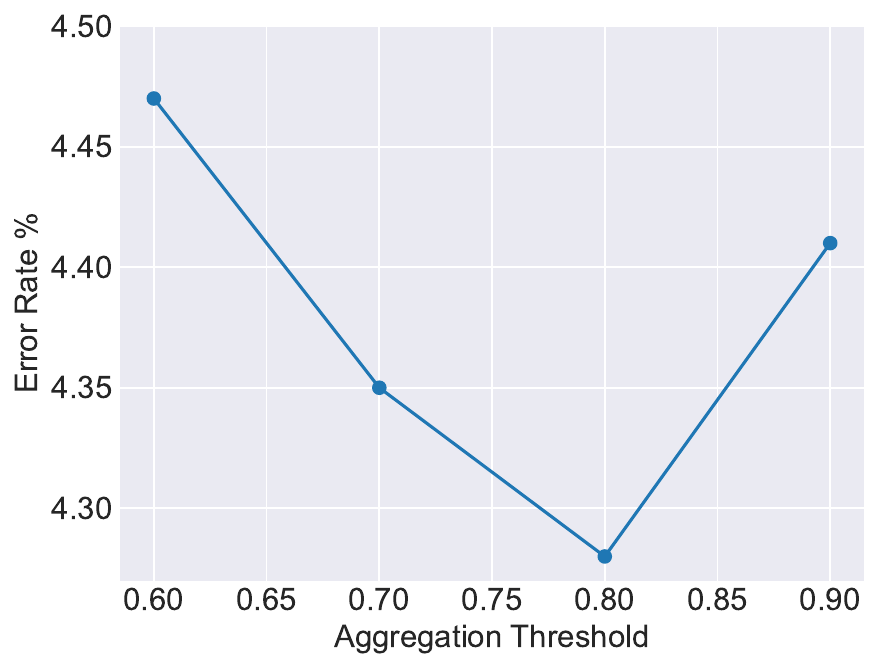}}
    \subcaptionbox{Number of Neighbors with $\tau_1=0.6$. \label{fig:nei_tau6}}{\includegraphics[width=.32\textwidth]{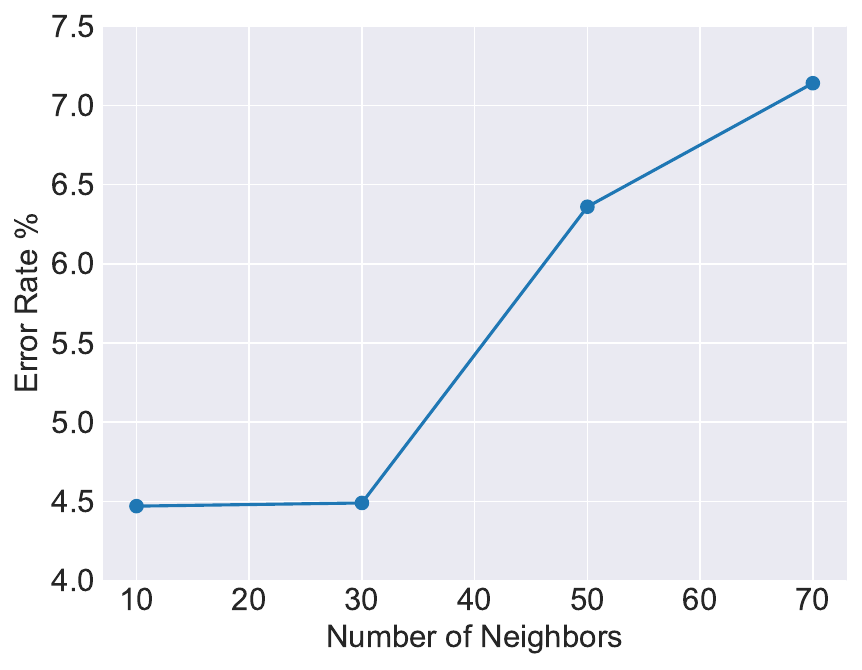}}
    
    \caption{Ablation study of the Second-level Interaction: (a)Error rate of the varying number of neighbors. (b)Error rate of the varying threshold of  aggregated pseudo-labeling. (c)Error rate of the varying number of neighbors with $\tau_1=0.6$. }
    \label{fig:ab_agg}
\end{figure}

    \noindent\textbf{Number of Neighbors.} Figure~\ref{fig:neighbor} illustrates the effect of different numbers of neighbors $K$ of Eq.~(\ref{eq:7}) on the aggregating pseudo-labeling. Note that $K=0$ means that DualMatch only uses Aligning Distribution. We observe that aggregating neighbor information improves model performance, but the number of neighbors within a scope has less influence on the model with a high confidence threshold. 

    \noindent\textbf{Aggregation Threshold.} We vary the threshold $\tau_1$ of Eq.~(\ref{eq:8}) to control the confident level of aggregated labels. Figure~\ref{fig:thershoud_1} shows the effect on aggregation threshold. When $\tau_1>0.6$, the aggregated labels are less affected by the unreliable pseudo-labels.

    We also investigate the effect of different aggregation thresholds combined with different numbers of neighbors on model performance. Figure~\ref{fig:nei_tau6} illustrates the effect of the number of neighbors $K$ with the aggregation threshold $\tau_1=0.6$. We observe that performance decreases as the number of neighbors increases with a low confidence threshold. The number of neighbors interacts with the aggregation threshold to ensure the reliability of aggregated pseudo-labeling.

\section{Conclusion}
    Our paper introduces a novel dual-interaction method for SSL that regulates diverse augmented representations with consistent class predictions and different class predictions with coherent feature representations. Leveraging this new perspective, we present a new SSL technique named DualMatch. DualMatch could learn more data-efficient representation and provide more robust pseudo-labels than the previous single-interaction-based SSL methods. Experimental results on both standard semi-supervised settings and more challenging class-imbalanced semi-supervised settings clearly demonstrate that DualMatch can achieve significant performance improvement.
\subsubsection{Acknowledgements} This work was supported in part by National Natural Science Foundation of China, Grant Number: 62206108 and Natural Science Foundation of Jilin Province, Grant Number: 20220101107JC.

\clearpage

\section*{Ethical Statement}
The purpose of this research paper is to explore a picture classification task under semi-supervised learning. In our study, we strictly adhere to ethical practice standards.

A number of ethical considerations were taken into account in conducting this study:
\begin{itemize}
    \item We ensure that no private data with others is directly involved in the study.
    \item We ensure that no indirect leakage of researcher or participant privacy occurs or privacy can be inferred in the course of the study.
    \item The data were collected from publicly available datasets. The data was analyzed using open source models. We ensure that the data is reliable and public, and that our analysis methods are widely accepted by the open source community and do not contain any bias or undue influence.
\end{itemize}

We also considered potential ethical issues that may arise in the course of the study. Semi-supervised learning has been widely used in various real-world scenarios, and this study explores the potential feature of data in semi-supervised scenarios, which is uninterpretable, and uses this feature to improve the performance and robustness of the model. With the development of deep learning, the potential features provided by the encoder may be interpreted, which may lead to the leakage of data privacy when improperly handled in the application of realistic scenarios.

We can conclude that our study is based on the open source community's code, models and public datasets. At this stage it does not cause privacy issues such as personal data leakage. Moreover, our study is currently not applied in real-world scenarios and there is no conflict of interest.

\clearpage
\bibliographystyle{splncs04}
\bibliography{mybibliography}

\end{document}